\documentclass{article}
\usepackage{ijcai25}

\usepackage{times}
\usepackage{soul}
\usepackage{url}
\usepackage[hidelinks]{hyperref}
\usepackage[utf8]{inputenc}
\usepackage[small]{caption}
\usepackage{graphicx}
\usepackage{amsmath}
\usepackage{amsthm}
\usepackage{booktabs}
\usepackage[switch]{lineno}
\usepackage{colortbl}

\usepackage[ruled]{algorithm}
\usepackage{algpseudocode}
\usepackage{multirow}
\usepackage{colortbl}
\usepackage[caption=false,font=normalsize,labelfont=sf,textfont=sf]{subfig}
\usepackage{graphicx}
\usepackage{tikz,xcolor,hyperref}
\usepackage{url}            
\usepackage{booktabs}       
\usepackage{amsfonts}       
\usepackage{nicefrac}       
\usepackage{microtype}      
\usepackage{array}
\usepackage{threeparttable}

\usepackage{amsmath}
\usepackage{bm}
\usepackage{amssymb}
\usepackage{framed}
\usepackage{pifont} 
\usepackage{textcomp}
\usepackage{stfloats}

\hypersetup{hidelinks,
	colorlinks=true,
	allcolors=black,
	pdfstartview=Fit,
	breaklinks=true}

\definecolor{mygray}{gray}{.9}
\definecolor{mypink}{rgb}{.99,.91,.95}
\definecolor{mycyan}{cmyk}{.3,0,0,0}

\title{Where and How to Enhance: Discovering Bit-Width Contribution for Mixed Precision Quantization}

\author{
Haidong Kang$^1$
\and
Lianbo Ma$^{1}$\footnote{Corresponding author}\and
Guo Yu$^2$\And
Shangce Gao$^3$\\
\affiliations
$^1$College of Software, Northeastern University\\
$^2$Institute of Intelligent Manufacturing, Nanjing Tech University\\
$^3$Faculty of Engineering, University of Toyama\\
\emails
\ hdkang@stumail.neu.edu.cn,
malb@swc.neu.edu.cn,
guo.yu@njtech.edu.cn,
gaosc@eng.u-toyama.ac.jp
}

\begin{document}

\maketitle

\begin{abstract}
 Mixed precision quantization (MPQ) is an effective quantization approach to achieve accuracy-complexity trade-off of neural network, through assigning different bit-widths to network activations and weights in each layer. The typical way of existing MPQ methods is to optimize quantization policies (i.e., bit-width allocation) in a gradient descent manner, termed as \underline{\textbf{D}}ifferentiable \underline{\textbf{MPQ}} (DMPQ). At the end of the search, the bit-width associated to the quantization parameters which has the largest value will be selected to form the final mixed precision quantization policy, with the implicit assumption that the values of quantization parameters reflect the operation contribution to the accuracy improvement. While much has been discussed about the MPQ’s improvement, the bit-width selection process has received little attention. We study this problem and argue that the magnitude of quantization parameters does not necessarily reflect the actual contribution of the bit-width to the task performance. Then, we propose a \underline{\textbf{S}}hapley-based \underline{\textbf{MPQ}} (SMPQ) method, which measures the bit-width operation’s direct contribution on the MPQ task. To reduce computation cost, a Monte Carlo sampling-based approximation strategy is proposed for Shapley computation. Extensive experiments on mainstream benchmarks demonstrate that our SMPQ consistently achieves state-of-the-art performance than gradient-based competitors.
\end{abstract}

\section{Introduction}
\label{sec:intro}
With the explosive growth in advanced IoT applications, it is challenging to deploy deep neural networks (DNNs) on resource-constrained devices (e.g., MCUs and tiny NPUs) \cite{kwon2023mobile,9954278,aggarwal2023chameleon}, which suffer from extremely limited memory (e.g., KB-level SRAM, and MB-level storage) and low computing speed \cite{wang2019haq,zheng2022energy}. This results in a big gap between the computational demands and limited resources. Therefore, it is desired to compress DNNs with no or minor performance degradation for efficient deployment. To achieve this goal, one typical way is to utilize the lower bit-width to quantize the entire network for lightweight and acceleration, a.k.a, fixed-precision quantization (FPQ) \cite{banner2019post,bai2023cimq}, where single-precision floating point weights or activations are mapped to lower bit-width ones. The recent developments in inference hardware have enabled variable bit-width arithmetic operations for DNNs, and this leads to the emergence of mixed precision quantization (MPQ) \cite{cai2020rethinking,sun2022entropy,ma2023ompq,dong2023emq,liu2023automatic,sun2024cim2pq}, which allows different bit-widths for different layers. Mixed precision of MPQ means a fine-grained bit-width allocation manner, where the quantization-insensitive layers can be quantized using much lower bit-widths than the quantization-sensitive layers. This way can naturally obtain more optimal accuracy-complexity trade-off than FPQ.

\begin{figure*}[t]
    \begin{center}
    \includegraphics[width=0.6\textwidth]{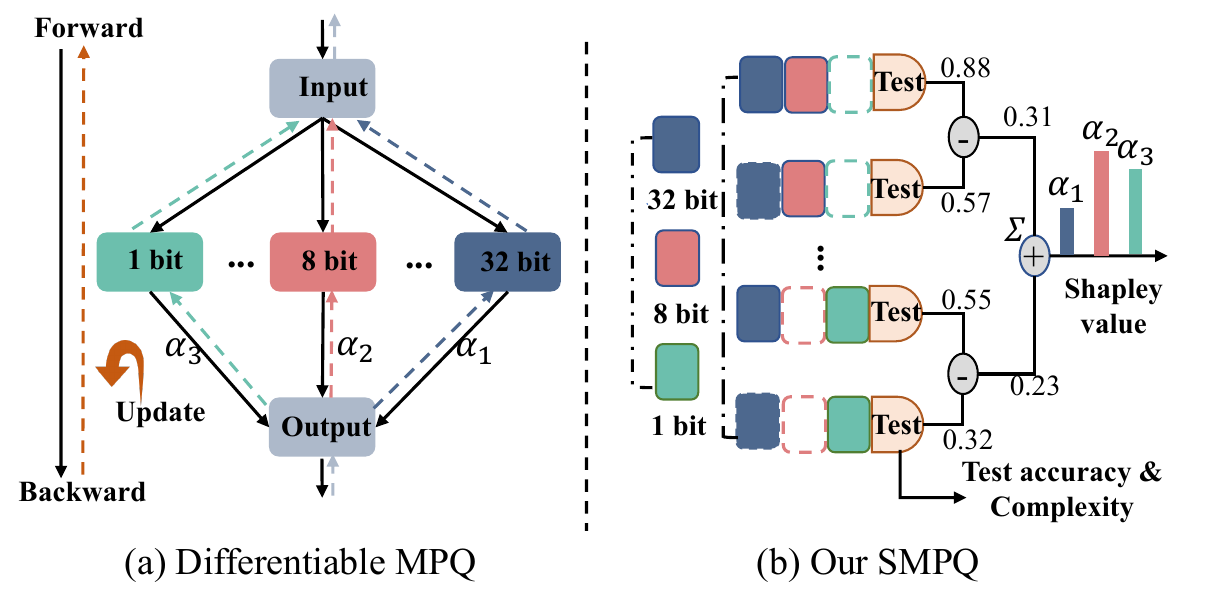}
    \end{center}
    \caption{The comparison between DMPQ and the proposed SMPQ. (a) DMPQ constructs a differentiable search space consisting of all bit-width candidates, and optimizes the learnable parameters $\alpha$ of bit-width updated by gradient descent, which does not reflect the contribution of bit-width candidates. (b) Our SMPQ directly evaluates the marginal contribution of bit-width candidates to the quantization task, according to the validation accuracy difference each possible bit-width candidate subset and its counterpart without the given bit-width candidate.}
    \label{motivation}
\label{fig:02}
\end{figure*}

\noindent{\textbf{Limitations of existing DMPQs.}}
As a typical way of existing MPQ methods, differentiable MPQ (DMPQ) aims to learn the optimal bit-width assignment in an end-to-end manner \cite{cai2020rethinking,zhang2021differentiable} (as shown in Fig. \ref{motivation}a). It applies continuous relaxation to transform the categorical choice of mixed precision quantization policy into continuous mixed precision quantization parameters \cite{yu2020search}. In this way, the DMPQ is relaxed to a differentiable quantization search problem, and the bit-width associated with the largest magnitude of learnable bit-width parameters (quantization parameters, $\alpha$) updated by gradient descent in each layer is selected to form the final mixed precision quantization policy. Such gradient-based bit-width selection process in DMPQ relies on an important assumption that \textit{the value of $\alpha$ updated by gradient descent represents the bit-width contribution to the accuracy improvement}. However, little work focuses on the validity of the above assumption of DMPQ.

In this paper, we attempt to understand and overcome this problem. At first, we find that the largest magnitude of learnable bit-width parameters $\alpha$ updated by gradient descent does not truly reflect the bit-width contribution in many cases, which degrades the performance of the derived mixed precision quantization policy.
In this work, this phenomenon is defined as \textbf{the $\alpha$’s pitfall issue}. Then, we observe that the bit-width operations in the quantization model are not independent of each other, which can be one potential reason of $\alpha$’s pitfall where the underlying relationships between bit-width operations are ignored.  

If the learnable bit-width parameters updated by gradient descent are not an excellent indicator (e.g., for optimization learnable bit-width parameters ($\alpha$)) of bit-width contribution, how to select optimal quantization policy for each layer during bit-width selection process? 
To tackle this issue, we propose a Shapley-based mixed precision quantization method, termed as SMPQ, which leverages Shapley value \cite{ancona2019explaining,castro2009polynomial,sun2024shapley} to attribute real contributions to players in cooperative game theory. Fig. \ref{motivation} shows the differences between our SMPQ and previous DMPQ methods. Shapley value directly measures the contributions of operations according to the validation accuracy difference. Meanwhile, it considers all possible combinations and quantifies the average marginal contribution to handle complex relationships between individual elements. Benefiting from these, Shapley value is effective for obtaining operation importance that is highly correlated with task performance.
Instead of relying on the learnable bit-width parameters updated by gradient descent, the bit-width selection of SMPQ considers the actual contributions of bit-width operations to quantization performance, as shown in Fig. \ref{motivation}b, which can be described by “Where” and “How” to search optimal quantization policies.

\noindent{\textbf{Contributions.}}  The main novelties of our work include:
\begin{itemize}
\item We discover that DMPQ suffers from the issue that the learnable bit-width parameters updated by gradient descent fail to reflect the actual bit-width contribution in many cases, and then we conduct deep analysis on the potential reason for such issue.  
\item  We propose a Shapley-based MPQ (SMPQ) method via capturing actual the contribution of bit-widths on the MPQ task of validation dataset. It is the first attempt in exploring enhanced bit-width selection of DMPQ.
\item  Extensive experimental results show that SMPQ consistently gets more optimal quantization policies than gradient-based counterparts. Furthermore, we find that SMPQ performs more efficiently than its comparators on resource-constrained devices. 
\end{itemize}

\section{Related Work and Preliminaries}
\subsection{Related Work}  
\label{sec:Revisit Mixed Precision Quantization}
Due to the page limit of the main text, the related work is provided in App. {A}.

\subsection{Rethinking the DMPQ Method}
\label{sec:Preliminaries}
\noindent{\textbf{Problem Setup}}. Suppose that a neural network $F$ consists of $n$ convolutional layers denoted as {$L_1, ..., L_n$}. Each layer $L_l$ ($1 \leq l \leq n$) has its corresponding set of weights $W_l$. 
The training process of the network $F$ is conducted by solving an Empirical Risk Minimization problem. To quantize weights $W_l$ and activations $A_l$, we define a search space $S^\alpha$ and $S^\beta$ with $n_w$ and $n_a$ bit-width candidates, respectively. To be specific, $S^\alpha$ represents the search space of weights, and $S^\beta$ denotes the search space of activations. To this end, the goal of MPQ is to find the optimal bit-width configuration for neural network $F$. Following the setting of method \cite{cai2020rethinking}, for layer $F_l$, the quantization function is typically defined as follows:
\begin{equation}
\small
	\centering
	\begin{aligned}
		\begin{matrix}Q_{l}(z)=\sum_{i=1}^{n_w}{o_{i}^\alpha W_l}\left(\sum_{j=1}^{n_a}{o_{j}^\beta A_l\left(z\right)}\right),&\\\mathrm{s.t.\ \ \ }\sum o_{i}^\alpha=1,\sum o_{j}^\beta=1,o^\alpha,o^\beta\in {0,1},&\\\end{matrix}
	\end{aligned}
	\label{eq1}
\end{equation}
where filter $f$ is parameterized by weight tensor $W^T_i$, and $z$ represents the filter input.
Our goal is to find the optimal bit-width configuration (${o_\alpha^\ast}$ and ${o_\beta^\ast}$) for the entire neural network $F$. Then, we can obtain:
\begin{equation}
\small
	\centering
	\begin{aligned}
		Y= \sum_{l=1}^{n} Q_{l}(F_l),
	\end{aligned}
	\label{eq2}
\end{equation}
where $Y$ represents output of the quantized network. However, we find that the search space is huge. For instance, in the case of ResNet-101, the number of possible bit-width configures reaches $6^{101}=3.9 \times 10^{78}$ for each input when there are only 6 bit-width candidates. Therefore, a key challenge for the MPQ problem is how to reduce the search time.

\noindent{\textbf{The DMPQ Method}}. To reduce search cost, a representative DMPQ method, EdMIPS \cite{cai2020rethinking} is proposed and developed, and it can consume less than 10 GPU hours in ImageNet1K.
The search space $B$ of DMPQ is represented by a supernet, denoted as G = (V, E), where each node $v_i$ $\in$ $V$ represents a latent representation, and each edge ${(i,j)}$ is associated with a bit-width $o^{(i,j)}$. The core idea of DMPQ is to transform the selection of discrete bit-width operation into a continuous optimization problem via continuous relaxation, and then optimize the supernet with gradient descent. The intermediate node is computed as a softmax mixture of candidate bit-widths:
\begin{equation}
\small
	\centering
	\begin{aligned}
	\bar{o}{(i,j)}(v_i)=\sum_{o\in\mathcal{S}}\frac{\exp(\alpha_o^{(i,j)})}{\sum_{o^{\prime}\in\mathcal{S}}\exp(\alpha_{o^{\prime}}^{(i,j)})}o(v_i),
		\end{aligned}
	\label{eq3}
\end{equation}
where $\alpha_o^{(i,j)}$ denotes the mixing weight of candidate bit-width $o^{(i,j)}$. $S$ represents the search space consist of $S^\alpha$ and $S^\beta$.
$\bar{o}$ is the input and mixed output of an edge. Given the optimal parameters $\alpha^\ast$, the mixed-precision network should be derived by discretizing the soft selector variables $\alpha_o^{(i,j)}$ into the binary selectors. With such relaxation, the DMPQ search can be performed by jointly optimizing the model weight $W$ and learnable bit-width parameters $\alpha$ for their corresponding bit-width candidate in a differentiable manner. In this work, we use a ``winner-take-all" method to determine which bit-width is selected and others removed, computed as follows:
\begin{equation}
\small
	\centering
	\begin{aligned}
	o_i^*=\left\{\begin{matrix}1,if~i~=~argmax_j~\alpha_o^{(i,j)},\\ 0,~otherwise.\end{matrix}\right.
		\end{aligned}
	\label{eq5}
\end{equation}
At the end of the search, the final mixed precision quantization policy is derived by selecting the maximum magnitude of learnable bit-width parameters $\alpha$ on every edge across all bit-width choices. Notably, our analysis does not limit the scope of EdMIPS, it also can be generalized to other differentiable MPQ (i.e., FracBits-SAT, GMPQ).

\section{The Magnitude-based Selection Pitfall Issue}
\label{sec:Preliminary}

\begin{figure}[t]
  \centering
 \subfloat[]{\includegraphics[width=3.1in]{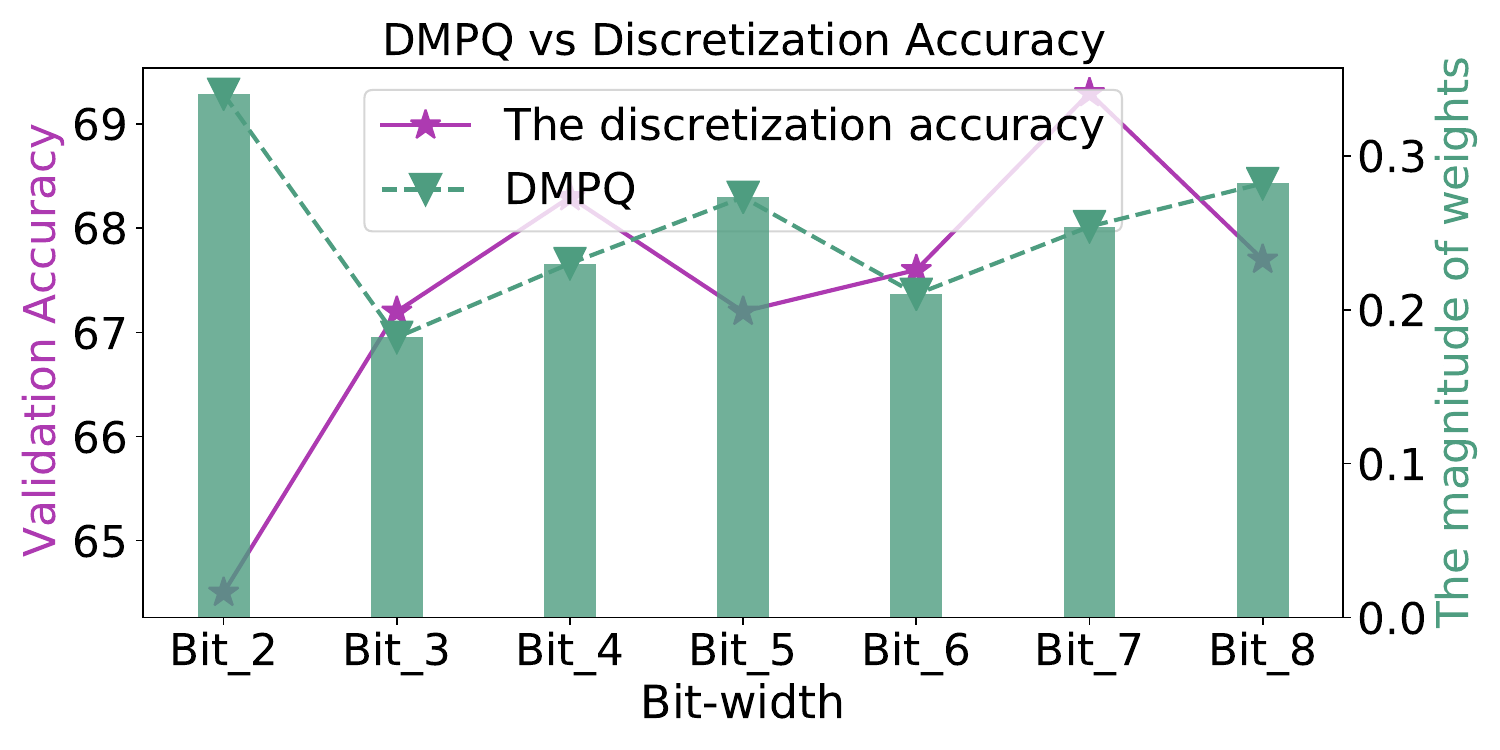}
    \label{fig:reliable_issue-a1}}
  \hfill
 \subfloat[]{\includegraphics[width=3.1in]{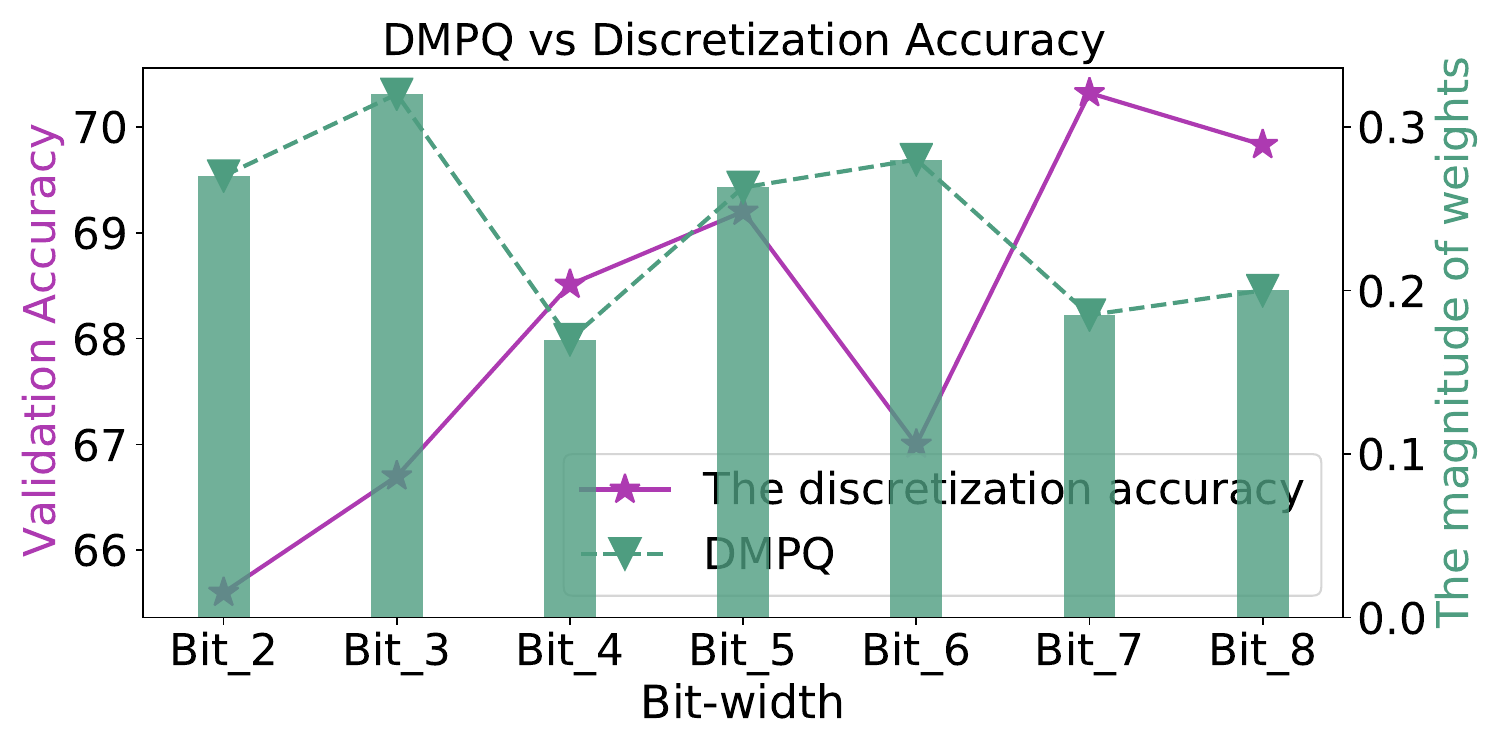}
    \label{fig:reliable_issue-b1}}
  \caption{ \textbf{DMPQ} \textit{v.s.} \textbf{discretization accuracy} on (a) Random edge 1 and (b) Random edge 2. To be specific, we randomly select 2 edges from a pre-trained EdMIPS supernet on the search space S1.
  }
  \label{fig:reliable_issue}
\end{figure}

\subsection{Rethinking Actual Contribution of Bit-width in DMPQ}
Despite the successful application of DMPQ in various model compression scenarios, it is demonstrated in the following that the gradient-based bit-width selection process of DMPQ does not accurately represent the actual contribution of the bit-width, i.e., the $\alpha$’s pitfall. As depicted in Eq. \ref{eq3}, the bit-width selection process of the gradient-based in DMPQ relies on an important assumption that the largest magnitude of the learnable bit-width parameters updated by gradient descent signifies the contribution of the bit-width, i.e., the bit-width that corresponds to the highest probability is used as the optimal quantization strategy. However, due to the lack of comprehensive theoretical support for the rationality of the learnable bit-width parameters updated by gradient descent, we argue that selecting the bit-width based on the learnable bit-width parameters updated by gradient descent does not truly reflect its contribution, which leads to the $\alpha$’s pitfall issue (i.e., the magnitude-based selection pitfall). In the following section, we will substantiate our hypothesis with empirical evidence.

\subsection{Experimentation on $\alpha$'s Pitfall of DMPQ}
\label{sec:3.4}
To verify the aforementioned issue, we first introduce search space S1 (consisting of weight bit \{2, 3, 4, 5, 6, 7, 8\} and activation bit \{2\}) to comprehensively explore relevant bit-width configurations and to facilitate the study of different aspects of quantization. Then, we conduct experimentation on learnable bit-width parameters updated by gradient descent does not represent actual bit-width contribution. 
This is achieved by calculating the discretization accuracy, which is obtained by using gradient descent to retrain quantized network\footnote{Such network is quantized via selecting a bit-width candidate for a layer while fixing involved bit-widths for other layers} from the search space S1. 
Fig. \ref{fig:reliable_issue} demonstrates that allocating small values of $\alpha$ updated by gradient descent to bit-widths can lead to high discretization accuracy at convergence.

These results indicate that the above $\alpha$'s pitfall issue results from the gradient descent process employed by DMPQ.
To reveal the reason behind the $\alpha$'s pitfall issue,
we argue that the $\alpha$ value from DMPQ is not always consistent with the corresponding contribution to the discretization accuracy in DMPQ. In fact, the contribution of bit-widths cannot solely be determined by the gradient's largest magnitude, and other factors such as the relationship of bit-width should be taken into account for accurate selection.

\subsection{Further Analysis}
\label{sec3.3}

Furthermore, we observe that the bit-widths in the quantization model are not independent but cooperative to each other during the learning of mixed precision quantization policy. Especially, as illustrated in Eq. \ref{eq1}, the differentiable learning strategy of DMPQ solely focuses on minimizing the loss function of learnable bit-width parameters ($\alpha$), but ignores the  effect of cooperation between different bit-widths on the quantization. In a sense, this can be one potential reason why DMPQ suffers from the issue of $\alpha$’s failure, where the underlying relationships between bit-widths are ignored.

\begin{figure}[t]
  \centering
    \includegraphics[width=0.85\linewidth]{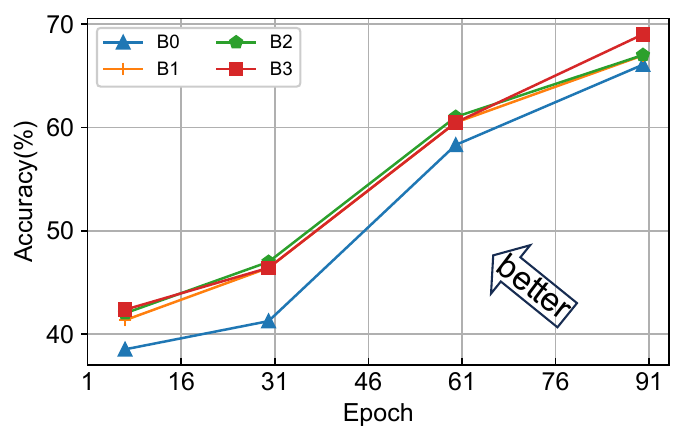}
  \caption{\footnotesize A study of bit-widths relationships. B0 denotes the final model quantized by DMPQ. B1 changes the bit-width of the third edge from 4 bits to 3 bits of B0, and B2 changes the bit-width of the fourth edge from 2 bits to 3 bits of B0. B3 alters the bit-widths of both edges of B0 by changing (4 bits, 2 bits) to (3 bits, 3 bits).}
  \label{fig:relationship}
\end{figure}

To validate our observation, motivated by method \cite{POQ}, we conduct further analysis to verify the issue, as shown in Fig. \ref{fig:relationship}. To demonstrate the underlying relationships between bit-widths on different edges, we propose the search space S2, which consists of weight bit \{1, 2, 3, 4\} and activation bit \{2, 3, 4\}. To be specific, B0 uses ResNet-18 (pre-trained in the ImageNet1K dataset) as the baseline model, and re-evaluates the discretization accuracy of the final mixed precision quantization
policy based on EdMIPS \cite{cai2020rethinking} method. 
B1 changes the bit-width of the third edge from 4 bits to 3 bits of B0, and B2 changes the bit-width of the fourth edge from 2 bits to 3 bits of B0. Moreover, B3 alters the bit-widths of both edges of B0 by changing (4 bits, 2 bits) to (3 bits, 3 bits). Finally, we re-evaluate B0, B1, B2, and B3 in the ImageNet1K dataset using the training settings shown in Table \ref{tab:hyperpara}.
As a result, we find the impacts of combinations of the two edges differ from the simple accumulation of their separate influence. Specifically, Fig. \ref{fig:relationship} shows that the accuracy of B3 at the convergence point exceeds the sum of B1 and B2, which indicates that there is a joint contribution between the 3th edge and the 4th edge for the whole model. That is, the impact of combinations of the two edges differs from the simple accumulation of their separate influence. This observation reveals the complex relationships between different bit-widths on different edges: some bit-widths can collaborate with each other, resulting in a significant joint contribution to the model's performance. Hence, we aim to resolve this issue in this paper by proposing a novel approach that leverages the Shapley value to estimate bit-width contribution from cooperative game theory, which can accurately reflect the contribution of the bit-width.

\begin{framed}
\noindent{\textbf{Conclusion.}} DMPQs indeed suffer from the issue that the learnable bit-width parameters updated by gradient descent fail to reflect the actual bit-width contribution.
\end{framed}

\section{Mixed Precision Search via Discovering Bit-width Contribution}
\label{sec:SMPQ}

\subsection{``Where to Search" in SMPQ}
\label{sec:where}

“Where to search”, i.e., where the optimal mixed precision quantization policy is derived, means defining the search space of MPQ. In this respect, we formulate the MPQ search space $B$ as a layer-wise supernet $G$ (consisting of bit-width sets $B$), similar to other DMPQ methods (as presented in Section \ref{sec:Preliminaries}). The only difference between them is that the updating of our supernet relies on the Shapley evaluation rather than the gradient in terms of $\alpha$, and thus requires no relaxation functions (e.g., $softmax$) that are typically used by other DMPQs (as shown in Eq. \ref{eq3}). One merit of the supernet is to provide weight sharing, which means that the weights of all quantization models sampled from the supernet directly inherit from those of the supernet without training from scratch. In this way, the search efficiency is greatly improved.

Let $\Omega(Q)$ be the complexity of quantized network through the precision policy $Q$. Then, the bit-width contribution weights $\alpha$ and network weights $W$ of supernet $G$ can be optimized in a layer-wise manner with the following objective:
\begin{equation}
\small
\centering
    \begin{aligned}
\min_{D_{val}} J_{v}(W^*,\alpha), \\
\text{s.t.}\quad\mathbf{W}^*=\arg\min_{D_{train}}\lim\limits_{\mathbf{W}}\left(\mathbf{W},\alpha\right) ,
&\Omega(Q)\leqslant\Omega_0. 
\end{aligned}
\label{eq4}
\end{equation}
where $J_{v}$ is validation loss, ${D_{train}}$ is training datasets, ${D_{val}}$ is validation datasets, and $\Omega_0$ is the resource constraint of the deployed device.


\subsection{``How to Search" in SMPQ}
\label{sec:How to Search}

“How to search” refers to how to optimize bit-width parameters $\alpha$ in supernet. Rather than the magnitude of bit-width parameters updated by gradient descent, we focus on their practical contribution to the target quantization performance. Moreover, we observe that the quantization operations in the supernet are often correlative with each other: the combinations of various bit-width quantizations may have different joint impacts on the quantization performance compared to their separate ones. To tackle such complex relationships, we employ the Shapley value to measure the average marginal contribution of the bit-width to the validation performance, where the MPQ search process is mapped into a cooperative game. The main procedures are as follows:

Assume that there are associated $N$ players in the game, and each subset of players $S \subseteq N$ (i.e., is coalition) is mapped to a real value $V (S)$ (as expected payoff of the players for cooperation) via a value function $V$. For the supernet, each layer has identical cell structures, and each cell has $| \mathcal{E} |$ edges each with $| \mathcal{O} |$ bit-widths. Then, the bit-width set, $N=\mathcal{O}\times\mathcal{E}={o^{(i,j)}}_{o\in\mathcal{O},(i,j)\in\mathcal{E}}$, can be regarded as players in the game, where all players cooperate towards the target performance $V (N )$. For bit-width $o^{\left(i,j\right)}$, its Shapley value $\psi_o^{\left(i,j\right)}$ is calculated by:

\begin{equation}
\tiny
	\centering
	\begin{aligned}
	\psi_{o}^{(i,j)}(V)=\frac{1}{|N|}\sum_{\mathsf{c}=\textit{N}|\circ(i,j)}\frac{V\left(S\cup\mathsf{o}^{(i,j)}\right)-V(S)}{\binom{|N|-1}{|S|}}.
		\end{aligned}
	\label{eqs1}
\end{equation}

In our scenario, we utilize the accuracy-complexity trade-off as the value function $V$ to evaluate the quantization performance. However, the direct computation of Eq. \ref{eqs1} needs to enumerate all possible subsets, and thereby requires $2^{\mid\mathcal{O}\mid\times\mid\mathcal{E}\mid}$ time complexity, which is computationally prohibitive. For efficient evaluation, we adopt Monte-Carlo sampling method \cite{castro2009polynomial} to get the approximation of the Shapley value, of which a truncated sample technique is also used to clip current sampling if the bit-width results in a significant performance drop. Accordingly, the search objective Eq. \ref{eq4} needs to be changed as:
\begin{equation}
	\centering
	\begin{aligned}
\alpha_o^{(i,j)}\propto\psi_{D_{val}}({J}_{v}(W^*,\alpha))~, \\
\text{s.t.}\quad\mathbf{W}^*=\arg\min_{D_{train}}\lim\limits_{\mathbf{W}}\left(\mathbf{W},\alpha\right) ,
&\Omega(Q)\leqslant\Omega_0, 
		\end{aligned}
	\label{eq6}
\end{equation}
where $\psi$ is the Shapley value of edge $(i,j)$. Since it is hard to directly solve the above objective, we need to update $\alpha_o^{(i,j)}$ using the Shapley value evaluated by the Monte-Carlo sampling-based approximate method:
\begin{equation}
	\centering
	\begin{aligned}
\alpha_{k}={\alpha_o^{(i,j)}}_{k-1}+\xi\cdot{\frac{q_{k}}{||q_{k}||_{2}}},  \\
\end{aligned}
	\label{eq7}
\end{equation}
where $\alpha_t$ is the contribution weight at the $k^{th}$ iteration during the search, ${q}_k$ denotes the accumulated Shapley value in the $k^{th}$ iteration, $\mid\mid\cdot\mid\mid_2$ is the L2 norm, and $\xi$ is the term coefficient. In the bi-level optimization, $\mathbf{W^t}$ is trained by gradient descent $\nabla L_{t}(w_{t-1},\alpha_{t-1})$ (as shown in Eq. \ref{eq6}) while $\alpha_{t}$ is optimized by Shapley value until convergence on training datasets. To alleviate redundant fluctuations induced by the sampling, we incorporate the momentum into the optimization for stabilization in the Monte-Carlo sampling process:

\begin{equation}
\footnotesize
    \centering
    \begin{aligned}
        \boldsymbol{q}_k=\beta\cdot\boldsymbol{q}_{k-1}+
        \lambda\cdot\frac{\boldsymbol{\psi}(Acc_{v}(\boldsymbol{w}_{k-1},\boldsymbol{\alpha}_{k-1}))}{||\boldsymbol{\psi}(Acc_{v}(\boldsymbol{w}_{k-1},\boldsymbol{\alpha}_{k-1}))||_2},
    \label{eq8}
    \end{aligned}
\end{equation}
where $\beta+\lambda=1$, the coefficients $\beta$ and $\lambda$ are used to balance the accumulated Shapley value and the current sampling, respectively. $Acc_{v}$ is the validation accuracy as value function, and $\boldsymbol{w}_{t-1}$ is the supernet weights at ${t-1}$ iteration. At the end of the search, the final precision policy is derived by selecting the bit-width with the largest contribution on each edge.

\subsection{Theoretical Analysis} 
\label{Theorical Analysis}
In this section, we analyze the expected error of SMPQ based on the Shapley theory \cite{ancona2019explaining,castro2009polynomial}  from a theoretical analysis perspective.

\noindent{\bf Theorem 1} \textit{(Upper-bounding of the risk on SMPQ}). \textit{Given the previous setting for SMPQ, for a set N of n players (bit-widths), according to Eq.7, let $\psi$ be the Shapley value of edge $(i, j)$, SMPQ is able to converge to the optimal point when:}

\begin{equation}
\footnotesize
\centering
\Delta_\psi\quad=\quad\sum_{i=1}^n50\times |(\min_{D_{val}} \psi({J}{v}(W^*,\alpha^{(k)}))|<\epsilon,
\label{appendix:eq1}
\end{equation}
where $\epsilon$ is the expectation error loss on specific datasets, and $abs(\cdot)$ is absolute function. We say that $\psi$ has converged to $\Delta_\psi$ and stopped the iterative process.

Assume that after the $k$th iteration, the weight coefficients are ${\alpha_o^{(i,j)}}_{k}$, and after the $(k+1)$th iteration, the weight coefficients are ${\alpha_o^{(i,j)}}_{(k+1)}$. To this end, we only need to show that $\psi({J}{v}(W^*,\alpha^{(k)}))$ will also converge. Therefore, we propose Lemma 1.

\noindent{\bf Lemma 1} \textit{(Bounding expectation $(E)$ of its marginal contribution).} According to Eq. \ref{eqs1}, the Shapley value $\psi_i(\cdot)$ of the game $\langle N, V\rangle$ for each subset of players (bit-widths) $S \subseteq N$ is the expectation $(E)$ of its \textit{\textbf{Marginal Contribution}} to players (bit-widths) that can be formulated as:

\begin{equation}
\footnotesize
\begin{aligned}
\centering
    \psi_i(N,V)=E[p_{o}^{(i,j)}(V)].
\label{appendix:eq6}
\end{aligned}
\end{equation}

The Shapley value can be explained as follows. Assume all the players are arranged in some order, with all orderings being equally likely. Afterward, $\varphi_i(N,v)$ is the anticipated marginal contribution (overall orderings), of player $i$ to the set of players who preceded him. According to the property of Shapley Value, for a mixed precision quantization game, one solution for $N$ players in the quantization game must exist with the biggest contribution $max(N)$ on validation datasets, which obtains best quantization weights for every player. In a mixed precision quantization game, each player (bit-width) only needs to maximize its own Shapley value (i.e., $\psi({J}_{v}(W^*,\alpha_o^{(i,j)}))$) so that $max(N)$ can be achieved as:

\begin{equation}
\footnotesize
\centering
    \begin{aligned}
\max_{D_{val}}V({N})
&=\sum_{i\in{N}}\max_{D_{val}}\psi(\mathbf{W}^*,\alpha_o^{(i,j)}).
\end{aligned}
\label{appendix:eq13}
\end{equation}

\noindent{\bf Proof of Lemma 1}. According to Eq. \ref{eqs1}, we can obtain:
\begin{equation}
\footnotesize
\centering
    \begin{aligned}
\psi_{o}^{(i,j)} = \mathbb{E}_{S \subseteq N \setminus \{o^{(i,j)}\}} \left[ V(S \cup \{o^{(i,j)}\}) - V(S) \right]=\\ \sum_{S \subseteq N \setminus \{o^{(i,j)}\}} \frac{|S|!(|N| - |S| - 1)!}{|N|!} \cdot \left[ V(S \cup \{o^{(i,j)}\}) - V(S) \right].
\end{aligned}
\label{th01}
\end{equation}

Now, we define marginal contribution of $o^{(i,j)}$ to $S$ as:
\begin{equation}
\footnotesize
\centering
    \begin{aligned}
p_{o}^{(i,j)}(V, S) := V(S \cup \{o^{(i,j)}\}) - V(S),
\end{aligned}
\label{th02}
\end{equation}
Based on Eq. \ref{th02}, we can get:
\begin{equation}
\footnotesize
\centering
    \begin{aligned}
\psi_{i}{(N, V)} = \mathbb{E}_{S} \left[ V(S \cup \{o^{(i,j)}\}) - V(S) \right] = E[p_{o}^{(i,j)}(V)].
\end{aligned}
\label{th03}
\end{equation}

According to the \textbf{``Efficiency"} of Shapley, we can obtain: 
\begin{equation}
\footnotesize
\centering
    \begin{aligned}
\max_{D_{val}}V({N})=\sum_{i\in{N}}\max_{D_{val}}\psi(\mathbf{W}^*,\alpha_o^{(i,j)}),
\end{aligned}
\label{th04}
\end{equation}
 The convergence of $\alpha_o^{(i,j)}$ relies on Eq. \ref{eq7}, \ref{eq8}, namely, $ \alpha_k = \alpha_{k-1} + \xi \cdot \frac{q_k}{\|q_k\|_2}$ , and $q_k = \beta q_{k-1} + \lambda \cdot \frac{\psi_{k-1}}{\|\psi_{k-1}\|_2}$ are exponential moving average, where $\psi_{k-1}=\psi(\text{Acc}_{val}(\mathbf{w}_{k-1},\alpha_{k-1}))$. If $\mathbf{W}$ converge in $D_{train}$, $\alpha_k$ and $q_k$ are stable in $D_{val}$. Therefore, we propose Lemma 2.



\noindent{\bf Lemma 2} \textit{(The convergence of quantized neural network).} According to \cite{Arora_Cohen_Golowich_Hu_2018}, we assume that quantized neural network is optimized by gradient descent with the learning rate $\eta$, depth $L$ of neural network, and deficiency margin $c > 0$. For any $\varrho > 0$, the loss of quantized neural network at iteration $T$ can be achieved as
follows:

\begin{equation}
\footnotesize
\centering
    \begin{aligned}
\frac{1}{\eta\cdot c^{2(L-1)/L}}\cdot\log\left(\frac{\ell(0)}{\varrho}\right)\le T.
\end{aligned}
\label{appendix:eq14}
\end{equation}

\noindent\textbf{Proof of Lemma 2.} Based on \cite{Arora_Cohen_Golowich_Hu_2018}, we have:
\begin{equation}
\footnotesize
\centering
    \begin{aligned}
\frac{1}{\eta \cdot c^{2(L-1)/L}} \cdot \log\left(\frac{\ell(0)}{\varrho}\right) \leq T,
\end{aligned}
\label{th05}
\end{equation}
Then, supernet weights \( \mathbf{W}^* =\arg\min_{D_{\text{train}}} \ell(\mathbf{W},\alpha) \) converge via gradient descent with learning rate \( \eta \), depth \( L \), and deficiency margin \( c > 0 \), for any \( \varrho > 0 \). The loss $L(\mathbf{W},\alpha)$ update by:
\begin{equation}
\footnotesize
\centering
    \begin{aligned}
\mathbf{w}_t = \mathbf{w}_{t-1} - \eta \nabla L(\mathbf{w}_{t-1},\alpha_{t-1}),
\end{aligned}
\label{th06}
\end{equation}
starting from initial loss \( \ell(0) \), the loss at iteration \( T \) satisfies:
\begin{equation}
\footnotesize
\centering
    \begin{aligned}
\ell(T) \leq \ell(0) \cdot \exp\left(-\eta \cdot c^{2(L-1)/L} \cdot T\right),
\end{aligned}
\label{th07}
\end{equation}
Then, we can derive:
\begin{equation}
\footnotesize
\centering
    \begin{aligned}
\exp\left(-\eta \cdot c^{2(L-1)/L} \cdot T\right) \leq \frac{\varrho}{\ell(0)},
\end{aligned}
\label{th08}
\end{equation}
Finally, we can obtain: 
\begin{equation}
\footnotesize
\centering
    \begin{aligned}
T \geq \frac{1}{\eta \cdot c^{2(L-1)/L}} \cdot \log\left(\frac{\ell(0)}{\varrho}\right).
\end{aligned}
\label{th09}
\end{equation}

As \( \mathbf{W} \to \mathbf{W}^* \), \( J_v(\mathbf{W}^*,\alpha^{(k)}) \) stabilizes, driving \( \psi \) to convergence.

\noindent\textbf{Proof of Theorem 1.} First, \( \alpha \) is updated using Shapley values.  By Lemma 2, for any \( \varrho > 0 \), we can derive
$\ell(W_T, \alpha) \leq \varrho, \quad \text{where} \quad T \geq \frac{1}{\eta \cdot e^{-2L}} \cdot \log \left( \frac{\ell(0)}{\varrho} \right).$ Thus, \( W_T \to W^* \). In bi-level optimization, \( W \) minimizes \( \ell(W, \alpha) \), while \( \alpha \) minimizes \( J_v(W, \alpha) \). By Lemma 1, the Shapley value is:
\begin{equation}
\footnotesize
\centering
    \begin{aligned}
\psi(N, V) = \mathbb{E}_{S} \left[ V(S \cup \{o^{(i,j)}\}) - V(S) \right].
\end{aligned}
\label{th10}
\end{equation}

In addition, the validation loss is bounded as follows:
\begin{equation}
\footnotesize
\centering
    \begin{aligned}
\min_{D_{\text{val}}} V(N) \leq \sum_{i \in \mathcal{N}} \max_{D_{\text{val}}} \psi(W^*, \alpha_o^{(i,j)}).
\end{aligned}
\label{th11}
\end{equation}

As \( \alpha_k \to \alpha^*\) (optimal quantization policy), we can derive \( \psi(J_v(W^*, \alpha^{(k)})) \) is stable. Then, we can derive:
\begin{equation}
\footnotesize
\centering
    \begin{aligned}
\min_{D_{\text{val}}} \psi(J_v(W^*, \alpha^{(k)})) \to 0.
\end{aligned}
\label{th12}
\end{equation}

Therefore, we have \( \Delta_\psi \to 0 \). For any \( \epsilon > 0 \), we can derive \( \Delta_\psi < \epsilon \). Theorem 1 is proven.

\begin{framed}
\noindent{\bf Remarks.} We further interpret how the above proofs can verify the effectiveness of our proposed SMPQ. First,
Lemma 1 proves that SMPQ can obtain bounding expectation of its marginal contribution for any subset of players/bit-widths, e.g., there exists an optimal contribution for any coalition, which proves there is a joint contribution between bit-widths on different edges (as presented in section \ref{sec3.3}). Such joint contribution of different bit-widths is a key to ensure high performance of our SMPQ. Second, Lemma 2 ensures the achieving of $\mathbf{W}^*$. Finally, theorem 1 shows that SMPQ can quickly converge towards a global minimum.
\end{framed}

\section{Experiments and Discussions}
\label{sec:Experiments and Discussions}

\begin{table}[t]
\centering
\resizebox{0.46\textwidth}{!}{
\begin{tabular}{c|cccccc}
\hline
\textbf{Network }            & \textbf{ResNet-18} & \textbf{MobileNetV2}   \\ 
\hline
Phase               & MPQ Training     & MPQ Training   \\
\hline
Epoch                             & 120                    &120          \\
Batch Size                         & 64                   &64          \\
Optimizer                              & Adam                   &AdamW        \\
Initial \textit{Lr}                 & 1e-3             & 1e-3  \\
\textit{Lr} Scheduler       & Cosine           & Cosine      \\
Weight Decay                           & -                  & -          \\
Warmup Epochs                         & -               & 30   \\
\hline
Random Crop              & \checkmark      &\checkmark        \\
Random Flip           & \checkmark      &\checkmark      \\
Color Jittering            &\checkmark            & -       \\
\hline
$\bm \xi$   & 0.1        & 0.1           \\
$\bm \beta, \lambda$     & {(0.8, 0.2)}        & {(0.8, 0.2)}       \\
Search Space ${S1}$    & {(2, 3, 4, 5, 6, 7, 8), (4)}&{(2, 3, 4, 5, 6, 7, 8), (4)}\\
Search Space ${S2}$    & {(1, 2, 3, 4), (2, 3, 4)}&{(1, 2, 3, 4), (2, 3, 4)}\\
Search Space ${S3}$    & [2, 8],[2, 8] &[2, 8],[2, 8]\\
\hline
\end{tabular}
}
\caption{Detailed hyper-parameters and training scheme of SMPQ for ResNet-18 and MobileNetV2 on ImageNet1K dataset.}
\label{tab:hyperpara}
\end{table}

\noindent\textbf{Experimental Settings.} 
Following previous quantization methods \cite{cai2020rethinking,chu2021mixed}, we train and evaluate our framework on single hardware platform, i.e., one NVIDIA Tesla A100 GPU with an Intel(R) Xeon(R) Gold 6133 CPU. In this paper, we use the training settings shown in Table \ref{tab:hyperpara}. Moreover, the training settings of ResNet-50, and Inception-V3 are the same as ResNet-18.

\noindent\textbf{Hardware.} All experiments are conducted on one NVIDIA Tesla A100 GPU with an Intel(R) Xeon(R) Gold 6133 CPU. Each experiment is executed on a single GPU at a time.



\begin{table*}[t]
\centering
\arrayrulecolor{black}
\resizebox{\linewidth}{!}{
\begin{tabular}{ccccccccc} 
\arrayrulecolor{black}\cline{1-8}
\multirow{2}{*}{Methods} & \multirow{2}{*}{Top-1 (\%) $\uparrow$} & \multirow{2}{*}{Search space.} & \multirow{2}{*}{Bit (W/A)} & \multirow{2}{*}{BOPs (G) $\downarrow$} & \multirow{2}{*}{Cost. $\downarrow$} & \multirow{2}{*}{Comp. $\uparrow$} & \multirow{2}{*}{Search method} &  \\
 &  &  &  &  &  & & \\ 
\cline{1-8}
Full-precision & 73.09 & & ~ & 1853.4 & ~ & ~ & ~ \\
\cline{1-8}
PACT \cite{choi2018pact} & 69.8 & \{5\}/\{5\} & 5/5 & 35.2 & - & 52.7$\times$ &Fixed-precision & ~ \\
LSQ \cite{bhalgat2020lsq+} & 67.6 & \{2\}/\{2\} & 2/2  & 23.1 & - & ~$80.2\times$ &Fixed-precision & ~ \\
PDQ \cite{chu2019mixed} & 65.0 & \{1,2,4,8\}/\{2\} & -/- & - & - &- & Fixed-precision & ~ \\
\cline{1-8}
Hybrid-Net* \cite{chakraborty2020constructing} & 62.7 & \{2\}/\{4\} & $\ast$/$\ast$ & - & - & - & Post-training Deterministic & ~ \\
HAWQ \cite{dong2019hawq} & 68.5 & \{2\}/\{4\} & $\ast$/$\ast$ & 34.0  & 15.6 & \cellcolor{cyan!30}54.5$\times$ & Sensitivity & ~ \\
EdMIPS \cite{cai2020rethinking} & 65.9 & [1,4]/[2,4] & $\ast$/$\ast$ & 34.7 & 9.5 &\cellcolor{cyan!30}54.5$\times$ & Differentiable & ~ \\
\textbf{SMPQ (ours)} & 68.70$\pm$0.05) & [1,4]/[2,4] & $\ast$/$\ast$ & \cellcolor{cyan!30}33.4 & \cellcolor{orange!30}2.4$\pm$0.08 &\cellcolor{orange!30}55.5$\times$ & Shapley-based & ~\\ 
\cline{1-8}
HAWQv3 \cite{yao2021hawq} & 70.4 & \{4\}/\{8\} & $\ast$/$\ast$ & 72.0  & -  &25.7$\times$ & Sensitivity & ~ \\
DNAS \cite{wu2018mixed} & 70.0 & \{1,2,4,8,32\}/\{1,2,4,8,32\} & -/- & 35.2  & -  &52.7$\times$ & Differentiable & ~ \\
SDQ \cite{SDQ} & 70.2 & - & 3.85/$3$ & \cellcolor{orange!30}25.1 & - &-& Differentiable & ~ \\
One-Shot \cite{One-Shot} & $<$70 & (2,4,8) & - & - & - &-& - & ~ \\
MetaMix \cite{MetaMix} & \cellcolor{orange!30}70.7 & - & 3.85/3 & - & - &-& - & ~ \\
\cline{1-8}
HAQ \cite{wang2019haq} & \cellcolor{cyan!30}70.4 & [2,8]/\{32\} & $\ast$/32 & 465  & -  &4.0$\times$ & RL-based & ~ \\
FracBits-SAT \cite{yang2021fracbits} & 70.6  & [2,8]/[2,8] & $\ast$/$\ast$ & 34.7 & - &53.4$\times$ & Differentiable & ~ \\
GMPQ \cite{chu2021mixed} & 69.9 & [2,8]/[2,8] & $\ast$/$\ast$ & \cellcolor{red!30}15.3 & \cellcolor{red!30}0.6 &\cellcolor{red!30}121.0$\times$ & Differentiable & ~ \\
\textbf{SMPQ (ours)} & \cellcolor{red!30}71.0$\pm$0.14 & [2, 8]/\{4\} & $\ast$/4 & 34.2 & \cellcolor{cyan!30}2.9$\pm$0.07 &54.2$\times$ & Shapley-based & ~\\ 
\textbf{SMPQ (ours)} & \cellcolor{red!30}72.6$\pm$0.03 & [2, 8]/\{8\} & $\ast$/8 & 59.7 & 4.9$\pm$0.02 &31.1$\times$ & Shapley-based & ~\\ 
\arrayrulecolor{black}\cline{1-8}
\end{tabular}
}
\caption{Accuracy and efficiency results for ResNet. The symbol ``Top-1" is the Top-1 accuracy of quantized model. The symbol ``Search space.” represents the search space. The symbol ``Bit (W/A)" denotes the average bit-width for weights and activation parameters. The symbol ``$\ast$" means mixed-precision quantization. The symbol ``Cost.” denotes the MPQ policy search time that is measured by GPU hours. The symbol ``Comp." means the compression ratio of BOPs. The symbol ``BOPs" denotes the bit operations.}
\label{tb:01}
\end{table*}

\begin{table*}[t]
\centering
\arrayrulecolor{black}
\resizebox{\linewidth}{!}{
\begin{tabular}{ccccccccc} 
\arrayrulecolor{black}\cline{1-8}
\multirow{2}{*}{Methods} & \multirow{2}{*}{Top-1 (\%) $\uparrow$} & \multirow{2}{*}{Search space.} & \multirow{2}{*}{Bit (W/A)} & \multirow{2}{*}{BOPs (G) $\downarrow$} & \multirow{2}{*}{Cost. $\downarrow$} & \multirow{2}{*}{Comp. $\uparrow$}  & \multirow{2}{*}{Search method} &  \\
 &  &  &  &  &  &  \\ 
\cline{1-8}
Full-precision & 72.5 & ~ &  ~ &337.9 & ~ & ~ \\
\cline{1-8}
PACT \cite{choi2018pact} & 61.4 & \{5\}/\{5\} & 4/4 & 7.42 & - & 45.5$\times$& Fixed-precision & ~ \\
LQ-net \cite{zhang2018lq} & 64.4 & \{4\}/\{4\} & 4/4  & 7.42 & - &45.5$\times$& Differentiable & ~ \\
RMSMP \cite{chang2021rmsmp} & 69.0 & - & $\ast$/$\ast$ & \cellcolor{orange!30}5.35  & -  &\cellcolor{orange!30}63.2$\times$& Sensitivity & ~ \\
\cline{1-8}
HAQ \cite{wang2019haq} & \cellcolor{cyan!30}71.5 & [2,8]/\{32\} & $\ast$/32 & 42.8  & 51.1  &7.9$\times$& RL-based & ~ \\
HMQ \cite{habi2020hmq} & 70.9 &  \{2,3,4,5,6,7,8\}/\{min(b(X),8)\} & $\ast$/$\ast$ & \cellcolor{red!30}5.32  & 33.5  &\cellcolor{red!30}63.5$\times$& Differentiable & ~ \\
FracBits-SAT \cite{yang2021fracbits} & \cellcolor{orange!30}71.6 & \{2,3,4,5,6,7,8\}/\{2,3,4,5,6,7,8\} & $\ast$/$\ast$ & \cellcolor{orange!30}5.35 & - &\cellcolor{orange!30}63.2$\times$& Differentiable & ~ \\
GMPQ \cite{chu2021mixed} & 70.4 & [2,8]/[2,8] & 3/$\ast$ & 7.4 & \cellcolor{red!30}2.6 &45.7$\times$& Differentiable & ~ \\
\textbf{SMPQ (ours)} & \cellcolor{red!30}71.7$\pm$0.01 & [2,8]/\{4\} & $\ast$/4 & \cellcolor{orange!30}5.35 & \cellcolor{orange!30}8.2$\pm$0.02 &\cellcolor{orange!30}63.2$\times$& Shapley-based & ~ \\ 
\textbf{SMPQ (ours)} & \cellcolor{red!30}72.0$\pm$0.02 & [2,8]/[2,8] & $\ast$/$\ast$ & \cellcolor{cyan!30}6.83 & \cellcolor{cyan!30}9.7$\pm$0.01 &\cellcolor{cyan!30}49.5$\times$& Shapley-based & ~ \\ 
\arrayrulecolor{black}\cline{1-8}
\end{tabular}
}
\caption{Results for MobileNetV2. The red, and orange, cyan indicate the best, second-best, and third-best, respectively. 
}
\label{tab:02}
\end{table*}

\begin{figure}[t]
  \centering
  \subfloat[]{\includegraphics[width=1in]{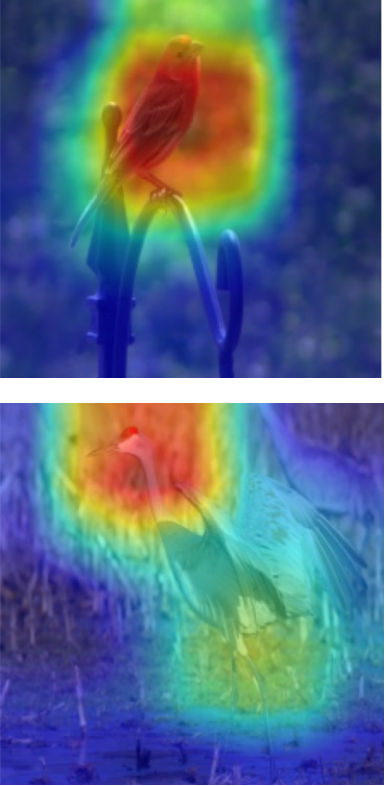}
    \label{fig:resnet32}}
    \hfill
  \subfloat[]{\includegraphics[width=0.98in]{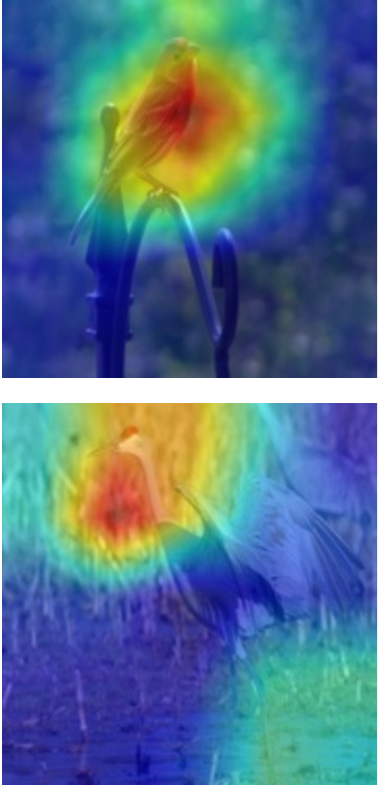}
    \label{fig:EdMIPS}}
  \hfill
\subfloat[]{\includegraphics[width=1in]{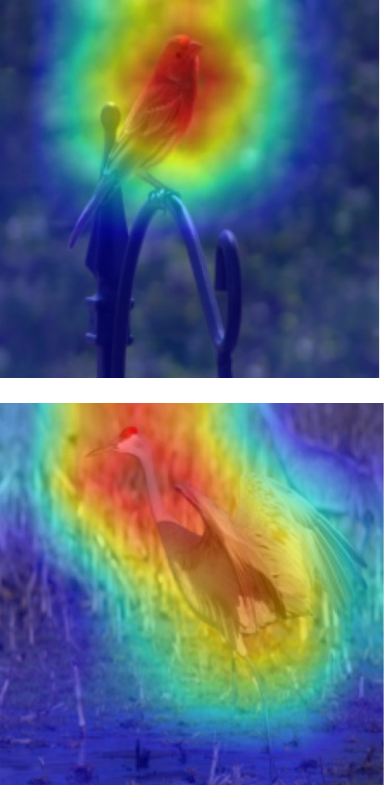}
    \label{fig:imagent_our}}
  \caption{The saliency maps (computed by \emph{Grad-cam} visualization on ImageNet1K. (a) the full precision ResNet-18, (b) MPQ policy searched by EdMIPS, and (c) MPQ policy searched by our SMPQ.}
  \label{motivation_tsn}
\end{figure}

\begin{figure}[t]
\centering
  \includegraphics[width=0.9\linewidth]{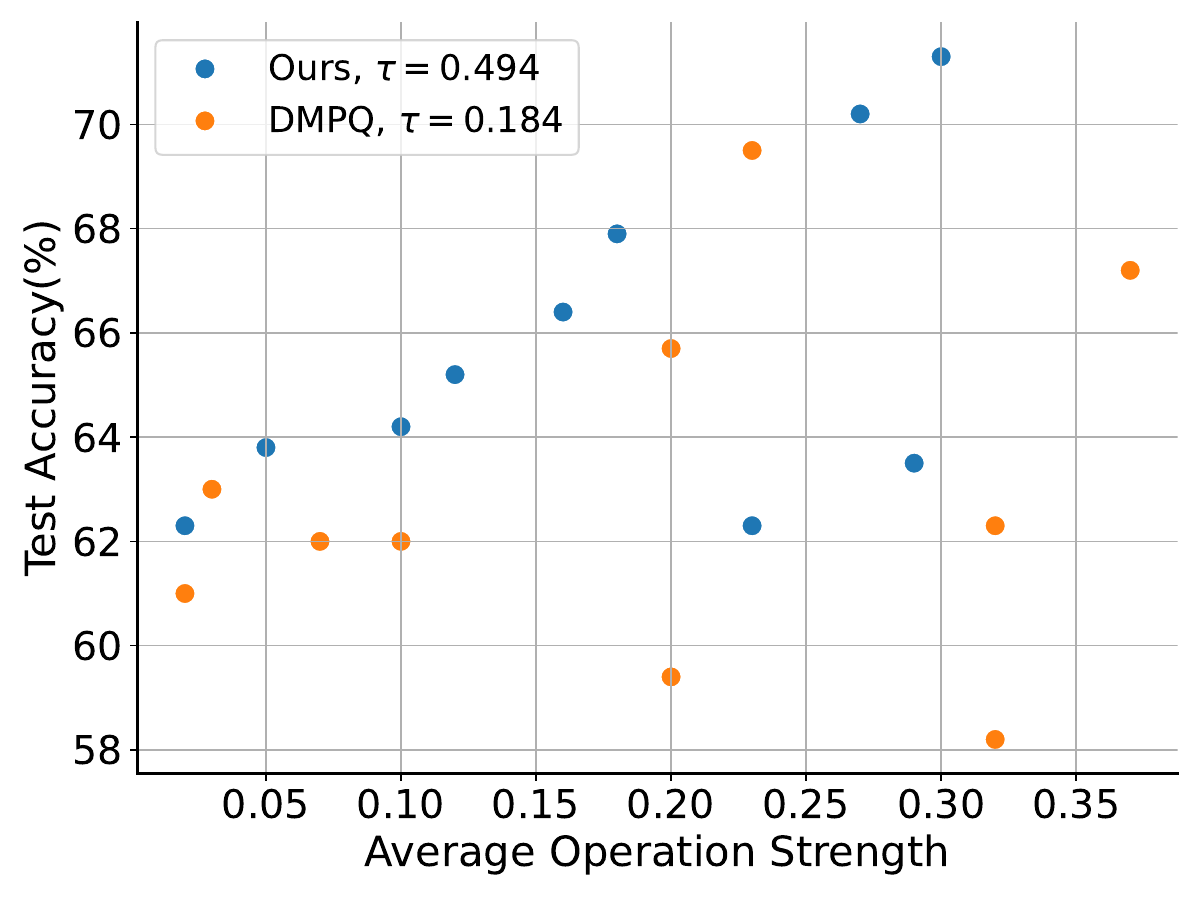}
  \caption{The correlation between test accuracy and average bit-width magnitude of 10 discrete quantization policies, which are sampled from the supernet using DMPQ and our SMPQ on ImageNet1K.}
  \label{robust}
\end{figure}

\begin{figure}[t]
  \centering
    \includegraphics[width=\linewidth]{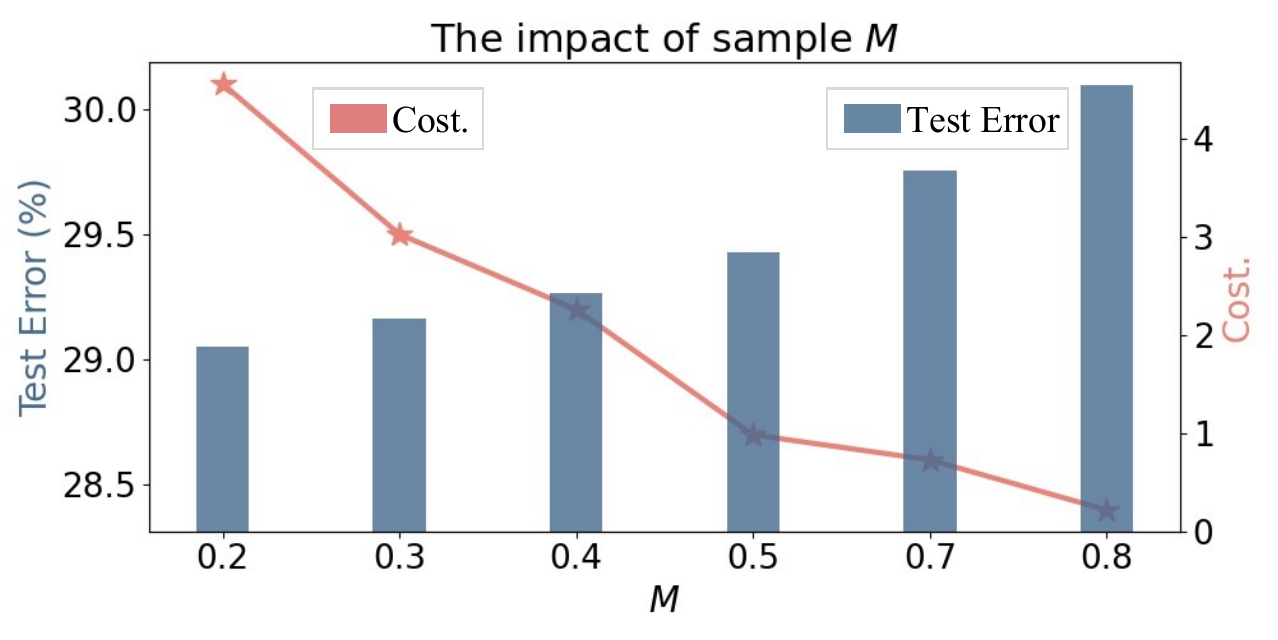}
   \caption{ The impact of samples $M$.
  }
  \label{landscaps}
\end{figure}

\begin{figure}[t]
  \centering
    \includegraphics[width=\linewidth]{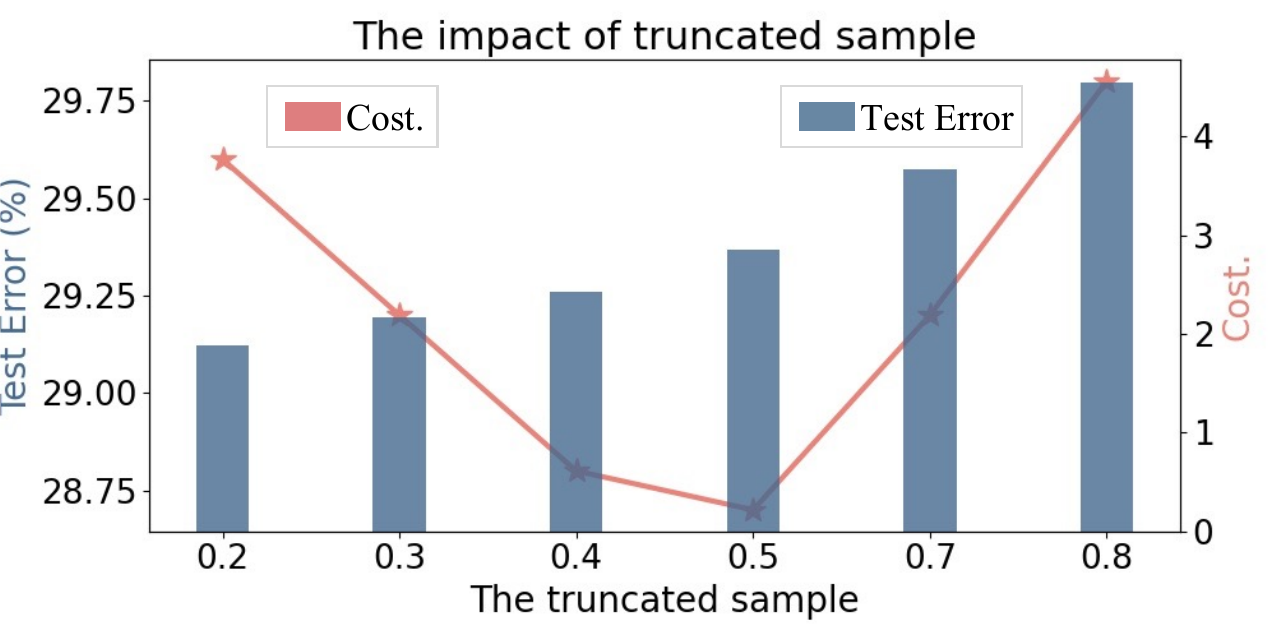}
  \caption{ The impact of the truncated sample.
  }
  \label{ab_t}
\end{figure}


\begin{table}[t]
    \centering
    \resizebox{\linewidth}{!}{
	\begin{tabular}{c|ccc|c}
		\hline
  {Method} & {$^{\#}$Param. (M) $\downarrow$} & {BOPs (G) $\downarrow$} & {Top1 $\uparrow$}  & {Bit (W/A)}\\
		\hline
		\multicolumn{5}{c}{ResNet-50}\\
		\hline
		Full-prec. & 97.8 & 3951.0 & 77.72\% &32/32 \\
  HAWQ \cite{dong2019hawq}  & 13.1 & 61.3 & 75.3\% &$\star$/$\star$ \\
		EdMIPS \cite{cai2020rethinking}   & 13.9 & 15.6 & 72.1\% &$\star$/$\star$ \\
  HAQ \cite{wang2019haq}  & 12.2 & 50.3 & 75.5\% &$\star$/$\star$ \\
  HMQ \cite{habi2020hmq}  & 15.6 & 37.7 & 75.5\% &$\star$/$\star$ \\
  BP-NAS \cite{yu2020search} & 11.2 & 33.2 & 75.7\% &$\star$/$\star$ \\
    EdMIPS-C \cite{cai2020rethinking}  & 13.7 & 16.0 & 65.6\% &$\star$/$\star$ \\
        \rowcolor{mygray}
  		SMPQ (ours) & 12.4 & 53.0  & 76.2$\pm$0.03\% &$\star$/$\star$ \\
    		\hline
		\multicolumn{5}{c}{Inception-V3}\\
		\hline
		Full-prec. & 90.9 & 5850.0 & 78.9\% &32/32 \\
        HWGQ \cite{cai2017deep}   & 29.4 & 376.2 & 71.0\% &$\star$/$\star$ \\
        Integer Only \cite{jacob2018quantization}   & 20.1 & 280.0 & 73.7\% &$\star$/$\star$ \\
		EdMIPS \cite{cai2020rethinking}   & 29.4 & 376.2 & 72.4\% &$\star$/$\star$ \\
        \rowcolor{mygray}
  		SMPQ (ours) & 19.6 & 265.0  & 74.6$\pm$0.14\% &$\star$/$\star$ \\
    \hline
	\end{tabular}}
    \caption{The performance comparison on larger networks.}
	\label{ex3}
\end{table}

\subsection{Comparison with State-of-the-Art Methods}
The experimental results (all models with the same parameter setting are averaged over 10 random repetitions) on ImageNet1K are presented in Tables \ref{tb:01}, and \ref{tab:02}, respectively. From the reported results, we can have the following observations:

\ding{182} As shown in Table \ref{tb:01}, our SMPQ obtains 6.7, 2.6, 2.0, and 1.9 percent improvements over EdMIPS, DNAS, FracBits-SAT, and MetaMix with 2.9$\pm$0.07 GPU hours for ResNet-18 on ImageNet1K, respectively. As shown in Table \ref{tab:02}, SMPQ achieves 72.0$\pm$0.02 accuracy, higher than peer competitors with 9.7$\pm$0.01 GPU hours for MobileNetV2. These results highlight the effectiveness of SMPQ. This is because that SMPQ methods can utilize the relationships between different bit-widths on different edges by our proposed SMPQ method, which can directly measure bit-width contribution to the performance of the quantized model on the validation sets.

\ding{183} When comparing the accuracy-complexity trade-off of SMPQ with baseline methods across various settings, it is evident that SMPQ provides a competitive accuracy-complexity trade-off under a variety of resource constraints. Importantly, this is achieved with significantly reduced BOPs (G). As shown in Table \ref{tb:01}, compared with the chosen baselines for ResNet-18, the compression ratio of SMPQ is 1.7$\times$ more than DNAS, 1.8$\times$ more than EdMIPS, and 7.8$\times$ more than HAQ. This shows the effectiveness of our SMPQ. 

\ding{184} \textbf{To the best of our knowledge, this work is the first to compare the performance among different search spaces}. As shown in Tables \ref{tb:01}, and \ref{tab:02}, with bit-width increasing, the performance of models further increases, implying that search space is a key factor to MPQ. By comparing with ``EdMIPS" under same search space, our SMPQ achieves competitive results with 68.7 top-1 accuracy and faster search speed, i.e., 2.4 GPU hours. v.s. 9.5 GPU hours. The main reason for better performance of SMPQ lies in that SMPQ can fully leverage relationship between different bit-widths through Shapley value. Shapley value is an excellent indicator of bit-width contribution, enabling accurately selects optimal mixed precision quantization policy for neural networks.

\subsection{Effectiveness for Larger Networks}
To demonstrate the generalizability of SMPQ, we conduct experiments that scale to larger networks (i.e., ResNet-50, Inception-V3, etc.). The results are listed in Table \ref{ex3}. As shown in Table  \ref{ex3}, compared with the chosen baselines for ResNet-50, the Top-1 accuracy of SMPQ is 4.1 more than EdMIPS, 0.7 more than HMQ, and 10.6 more than EdMIPS-C. By comparing with the chosen baselines for Inception-V3, the Top-1 accuracy of SMPQ is 3.6 more than HWGQ, 0.9 more than ``Integer Only", and 2.2 more than EdMIPS. We can figure out that our approach is far better than peer competitors, which implies good generalizability on larger networks.

\begin{table}[htbp]
    \centering
    \resizebox{\linewidth}{!}{
    \begin{threeparttable}{
\begin{tabular}{c|cc|cc|cc|cc}
\hline
     
\multirow{2}*{\textbf{step size $\xi$}} & \multicolumn{2}{c|}{\textbf{$\beta=0.25$}}& \multicolumn{2}{c|}{\textbf{$\beta=0.5$}} & \multicolumn{2}{c|}{\textbf{$\beta=0.75$}}& \multicolumn{2}{c}{\textbf{$\beta=1$}}
  \\ \cline{2-9}
  &\textbf{Err.(\%)} & \textbf{BOPs.(G)}  &\textbf{Err.(\%)} & \textbf{BOPs.(G)}
  &\textbf{Err.(\%)} & \textbf{BOPs.(G)} &\textbf{Err.(\%)} & \textbf{BOPs.(G)}  \\\hline
             0.01 &$30.5 $&$31.5 $&30.2 &$31.7 $
             &$29.9 $ &$33.6 $ &$29.4 $ &$33.1 $   \\
             0.05 &$31.1 $&$30.7 $&30.4 &$ 31.4$
             &$29.0 $ &$34.2 $ &$29.5 $ &$34.6 $   \\
            0.1 &$29.4 $&$34.6 $&29.7 &$34.8 $
             &$29.0 $ &$34.2 $ &$29.5 $ &$34.6 $   \\
            0.5 &$ 31.4 $&$32.5 $&31.6 &$33.7 $
             &$29.5 $ &$35.1 $ &$30.1 $ &$32.6 $   \\ \hline        
\end{tabular}
	 }
\end{threeparttable}
	}
    \caption{Ablation studies of $\beta$ and $\xi$ for ResNet-18, where $\beta=1-\lambda$.}
       
    \label{tab:abl2}
\end{table}

\subsection{Correlation Analysis}
Then, we perform a correlation analysis using Kendall's Tau ($\tau$) coefficient. After the search phase, we select 10 discrete quantization policies from the supernet and calculate their corresponding bit-width strength by averaging the magnitude of the learnable bit-width parameters. We then plot the test accuracy \textit{vs.} average bit-width magnitude obtained by DMPQ and our SMPQ.
To measure the correlation between test accuracy and average bit-width magnitude, the results of Kendall's Tau (KTau) coefficient on ImageNet1K are depicted in Fig. \ref{robust}.
The larger the KTau value, the more predicted contribution matches the real contribution.
From Fig. \ref{robust}, we can see that our SMPQ achieves a higher correlation with the test accuracy ($\tau$ = 0.494), while DMPQ (the learnable bit-width parameters are updated by gradient descent) shows a worse correlation with the final test accuracy.
In summary, the correlation analysis demonstrates that our proposed SMPQ can predict the real contribution to bit-widths more accurately than DMPQ.

\subsection{Ablation Study}
\label{sec:ablation}

\noindent{\textbf{Influence of  sampling $M$ and truncated sample.}} In fact, the two factors are crucial for model accuracy and search cost, as shown in  Fig. \ref{landscaps} and \ref{ab_t}. With the increase of $M$, the test error degrades significantly, while the search cost increases dramatically. As a result, when $M$ = 10, we obtain accuracy-complexity trade-offs for SMPQ. For truncated sample, medium truncated sample (threshold=0.5) achieves the best accuracy-complexity trade-off.

\noindent{\textbf{Influence of $\beta$ and $\xi$.}} Then, by varying  $\beta$ and $\xi$, we evaluate their influence with respect to the model accuracy. As shown in Table \ref{tab:abl2},
we can find that our SMPQ is insensitive to $\beta$ and $\xi$, and when $\beta=0.75$ and $\xi=0.05$, we get the optimal results.

\section{Conclusion}
\label{conclusion}
In this paper, we attempt to understand and enhance MPQ methods from the bit-width selection perspective. We examine the magnitude-based quantization selection process of DMPQ and provide empirical evidence to show the $\alpha$’s pitfall issue, i.e., the value of $\alpha$ cannot well reflect the actual bit-width contribution, and then conduct deep analysis on the potential reason of such issue. To overcome this issue, we propose a Shapley-based MPQ method that directly estimates the bit-width contribution via its contribution to the performance of the quantized model on the validation sets. Specifically, the Shapley value of bit-widths can be efficiently approximated by a Monte-Carlo sampling algorithm with early truncation. The proposed method is able to consistently obtain more optimal quantization policies than magnitude-based MPQs on a set of datasets and several search spaces. 



\section*{Acknowledgments}
This work is supported by National Natural Science Foundation of China under Grant 62472079, Fundamental Research Funds for the Central Universities (N2417008, N2417003), the Key Technologies R$\&$D Program of Liaoning Province (2023JH1/10400082), and Guangdong Basic and Applied Basic Research Foundation (2024A1515012016).

\bibliographystyle{named}
\bibliography{ijcai}

\end{document}